\documentclass[10pt,twocolumn,letterpaper]{article}

\usepackage{cvpr}              

\usepackage{graphicx}
\usepackage{amsmath}
\usepackage{amssymb}
\usepackage{booktabs}

\usepackage[pagebackref,breaklinks,colorlinks]{hyperref}

\usepackage[capitalize]{cleveref}
\crefname{section}{Sec.}{Secs.}
\Crefname{section}{Section}{Sections}
\Crefname{table}{Table}{Tables}
\crefname{table}{Tab.}{Tabs.}

\def\cvprPaperID{*****} 
\def\confName{CVPR}
\def\confYear{2022}

\begin{document}

\title{Pattern Based Multivariable Regression using Deep Learning (PBMR-DP) \confName~Proceedings}

\author{Jiztom Kavalakkatt Francis\\
Department of Computer Engineering\\
Iowa State University\\
{\tt\small jiztom@iastate.edu}
\and
Chandan Kumar\\
Department of Computer Science\\
Iowa State University\\
{\tt\small chandan@iastate.edu}
\and
Jansel Herrera-Gerena\\
Department of Computer Science\\
Iowa State University\\
{\tt\small janselh@iastate.edu}
\and
Kundan Kumar\\
Department of Computer Science\\
Iowa State University \\
{\tt\small kkumar@iastate.edu}
\and
Matthew J Darr\\
Department of Agricultural and BioSystems Engineering\\
Iowa State University\\
{\tt\small darr@iastate.edu}
}

\maketitle

\begin{abstract}
    We propose a deep learning methodology for multivariable regression that is based on pattern recognition that triggers fast learning over sensor data. We used a conversion of sensors-to-image which enables us to take advantage of Computer Vision architectures and training processes. In addition to this data preparation methodology, we explore the use of state-of-the-art architectures to generate regression outputs to predict agricultural crop continuous yield information. Finally, we compare with some of the top models reported in MLCAS2021.
    We found that using a straightforward training process, we were able to accomplish an MAE of 4.394, RMSE of 5.945, and $R^2$ of 0.861. 

\end{abstract}

\section{Introduction}
\label{sec:intro}
In the recent years, machine learning algorithms have been improving dramatically in different areas. Unsupervised methods have been incorporated in the deep learning field to solve image-based problems, sound, and text. We also notice that neural network architectures have changed and consequently, they have changed the training process. Some works have also tried to make changes into the backbone network \cite{vaddi_kim_kumar_shad_jannesari_2021} to achieve better results. But sometimes, the innovation blinds some improvement in promising ideas that were not developed to a higher potential. Here, we present our work that combines state-of-the-art image architecture and regression.

Inspired by the data provided in \cite{2021}, a sensor dataset containing information of multiple sensors with time-stamp. We decided to take a different approach and explore the conversion of this dataset into images (\Cref{subsec:Input_data}). This conversion opens the doors of Computer Vision (CV) models for tabular data. First, we explored the conversion of sensor data into an accurate image-like data, and then make changes in the neural network architecture as common CV architectures do not tend to give regression as output which was the case for our model. This allows us to perform multivariable regression as in \cite{multi} which is pattern-driven instead of data-driven.

\begin{figure}[t]
    \centering
    \includegraphics[scale=0.34]{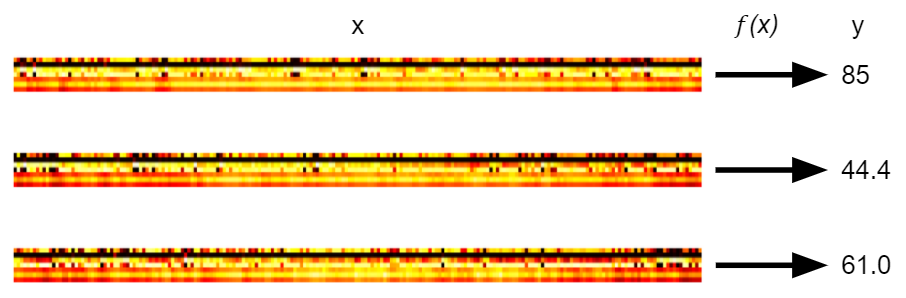}
    \caption{Three samples of how the tabular input data ($x$) looks when converted to an image mapped, by our model $f(x)$ , to their predicted ($y$) value.}
    \label{fig:Data_visualzation}
\end{figure}

\subsection{Contribution}
\begin{figure*}[t]
    \centering
    \includegraphics[scale=0.45]{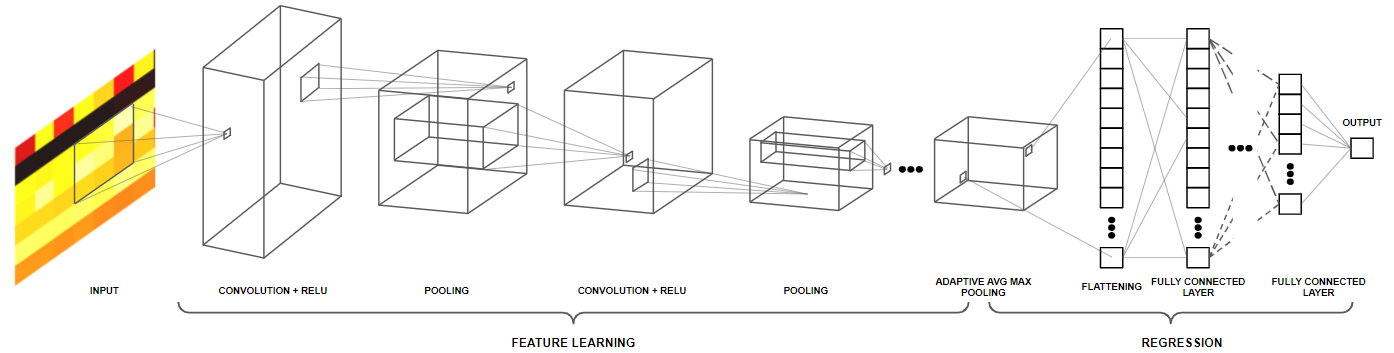}
    \caption{Here we present our proposed model architecture. The input is the pre-processed image like 3D-array passed to Convolutional Neural Network (Feature Learning). The generated output is adjusted using an Adaptive Concat Pooling mechanism and Fully Connected Network (FCN) to finally generate the required single regressor output.}
    \label{fig:ModelArchitecture}
\end{figure*}

In this work, we present two major contributions. The first one is constructing sensors-to-image conversion in which tabular data can be represented as an image. This facilitates the use of modern CV architectures. Secondly, using these sensors-to-image samples to predict continuous crop yield values. 

\section{Related Works}
\label{sec:related_works}
We did not want to base our architecture on long short-term memory (LSTM), which usually takes a lot of resources to perform the training process and hence compelled us towards using images. This led us to do exploration over methods that involved images and regression. To get started, we explored the idea around image age detector, which affirmed our concerns. Work done in \cite{Nada2020AgeAG}  deals with the creation of two Convolutional Neural Networks (CNNs), one to predict gender and another for age prediction with a classifier instead of a regressor. In practice, there is not much done in terms of having a regression output from an image-based model. 

Finding that many approaches to what, in our knowledge, are regression problems have in common the characteristics of converting it to a classification problem led us to explore other fields. We started by looking at \cite{Fischer2015ImageOE}, in which they work on a network able to predict the rotation angle of given images. A similar idea can be seen in \cite{mahendran20173d}, which shows a CNN regression framework for predicting 3D pose estimation. 

In another hand, we explore the conversion of sensor data into images such as \cite{s20010168}. The data was also serialized in such work and represented different factors that we did not deal with. Therefore, their conversion was more complex than in this work, but the idea of generating these images is viable. The melspectogram generates images using the Librosa\cite{Mcfee2015} package, allowing classification of sounds based on patterns. Visualizing sound as an Image \cite{9034644, becker2019interpreting} with DNNs improves accuracy and reduces computational requirements from classical methods of event or pattern recognition \cite{Kons2013AudioEC}.  Proving that the translation from another medium to image has worked in the past. 

The use of CNNs in image classification has become the day's standard. The image classification revolution began with the use of AlexNet \cite{NIPS2012_c399862d}. The inception models are carefully customized multi-branch architectures with carefully designed branches. ResNet \cite{7780459}, ResNeXt \cite{xie2017aggregated}, and EfficientNet \cite{tan2020efficientnet} are some examples of modern architectures.


Time series data becomes complex when the number of sensors and the frequency of data recording increases. The current solution is regression to find the best fit based on the multivariable data. Early proposed solutions require the conversion and generation of custom CNN like a 2 stage CNN proposed in \cite{ijcai2019-932}. The usage of detecting patterns requires much pre-processing with feature engineering. The process is time-consuming and will require extensive study of the correlation of each input date with the training data.

\begin{table*}[t]
\centering
\begin{tabular}{|p{3cm}|c|c|c|c|c|c|c|c|c|}
    \hline
    & \multicolumn{9}{c|}{\textbf{Performance}} \\
    \cline{2-10}
    \textbf{Models}& \multicolumn{3}{c|}{\textbf{MAE$\downarrow$}} & \multicolumn{3}{c|}{\textbf{RMSE$\downarrow$}} & \multicolumn{3}{c|}{\textbf{$R^2\uparrow$}}\\
    \cline{2-10}
    & \textbf{SGD} & \textbf{Adam} & \textbf{LARS} & \textbf{SGD} & \textbf{Adam} & \textbf{LARS} & \textbf{SGD} & \textbf{Adam} & \textbf{LARS}\\
    \hline
    \hline
    ResNet 50 & 4.529 & 5.496 & 4.644 & 5.963 & 7.258 & 6.266 & 0.849 & 0.792 & 0.845 \\ \hline
    EffificientNet B0 & 5.535 & 5.232 & 6.577 & 7.312 & 6.958 & 8.586 & 0.789 & 0.809 & 0.709 \\ 
    \hline
    ResNeXt50 & \textbf{4.394} & 5.371 & 5.191& \textbf{5.945} & 7.118 & 6.889 & \textbf{0.861} & 0.799 & 0.812  \\ \hline
  \end{tabular}
  
  \caption{Performance metrics with different standard models using different Optimizers. All models run with the learning rate and batch size specified in \Cref{sec:experiment}.}
  \label{tab:CNN_Multi_model_metric}
\end{table*}

\begin{table*}
    \centering
    \begin{tabular}{|c|c|c|c|c|}
    \hline
    {\textbf{Competition Teams}} & {\textbf{Model approaches}} & \multicolumn{3}{c|}{\textbf{Performance}}  \\
    \cline{3-5}
     & &\textbf{MAE$\downarrow$} & \textbf{RMSE$\downarrow$} & \textbf{$R^2\uparrow$} \\
     \hline
     \hline
    QU(exp006) & Statistical Modelling  & 4.41 & \textbf{5.89} & \textbf{0.87} \\
     \hline
    CUFE & ensemble Regression  & 4.42 & 5.95 & 0.86\\
     \hline
     Star & M/4* 1D CNN with Ensemble  & 4.47 & 5.95 & 0.86\\
     \hline
     Elendil & M/7 * 1D CNN with Ensemble 5 & 4.47 & 5.95 & 0.86\\
     \hline
     AA2 & XgBoost  & 4.6 & 6.15 & 0.85 \\
     \hline
     PBMR-DP & ResNeXt 50 - SGD & \textbf{4.39} & 5.94 & 0.86 \\
     \hline
    \end{tabular}
    \caption{Comparison with the models submitted in MLCAS2021 Challenge using the same evaluation metrics.}
    
    \label{tab:competiion_metrics}
\end{table*}

\section{Method}
In this section, we will explore the input pipeline, architecture design, and our approach to utilize the feature learning ability of DNNs to solve multivariable regression problems.


\subsection{Input Data}
\label{subsec:Input_data}
Our dataset is based on temporal data, which is computed in real-time. It can be noisy due to the different measuring speeds of the dataloggers \cite{jiztom2019} or the sensors' measurement of the values themselves. The initial assumption is that all the data is measured over the same time-space, corrected, or spread to a fixed tabular form. Sensor data, in particular, is considered as the ranges for sensors are absolute, ensuring that on normalization stage in pre-processing values are between 0 and 1.

The Soybean Crop Yield dataset found in the MLCAS2021 challenge is composed of  93000 samples over 214 days (1 crop season) with seven sensor readings, each pointing to a Single Crop Yield (y). There is also some additional information such as genotype ID, location ID, and year for each sample. This additional information is also normalized and treated like a sensor. Therefore, it is used as one of the rows in the input data after pre-processing.

\subsection{Pre-processing}
\label{subsec:preprocessing}
Before feeding machine learning models with data, we must pre-process the original data and statistically analyze it extensively before using them as input data. This process is time-consuming and requires human and computer resources to verify the correlation of the data to the output it is being trained with. Our process is different since we convert tabular data into images. The input data is arranged in the sensor data format as rows with time along the y-axis. Unlike most image processing steps in CNNs, we apply a Row Normalization technique. Each row is normalized based on the absolute range of the sensors \cref{Equ:Normilation}. This makes sure the final table generated contains values between 0 and 1.

\begin{equation}
    \overrightarrow{x_{ij}} = \frac{x_{ij} -\sigma(s_i)}{\lambda(s_i)-\sigma(s_i)}
    \label{Equ:Normilation}
\end{equation}
where $\overrightarrow{\boldsymbol{x}_{ij}} \in [0,1]$  is the normalized data point at positions $i, j$. The values in $x_{ij}$ represent the original tabular data in which $i$ represents the row (our sensor), and $j$ the time in our dataset. In addition, $\sigma(s_i)$ and $\lambda(s_i)$ represent absolute minimum and maximum values of sensor $s_i \in S$ where $S$ is the set of all the sensors. 

Our data preparation method from tabular data explained above allows it to be fed directly to CNNs without major modifications to the architecture. The tabular data must be across a common measurement axis, such as time series or measured at the same interval. If any values are missing in the tabular data, we will use the immediate past data to fill the missing blank in the table. This property of time series data helps ensure noise is reduced to a minimum in the input data. The generated tabular data is normalized row-wise based on the absolute range of the measured variable (sensor). \cref{fig:Data_visualzation} shows how the data can be visualized with patterns.
\begin{table*}[t]
    \centering
    \begin{tabular}{|c|c|c|c|}
    \hline
    {\textbf{Regression Analysis Techniques}} & \multicolumn{3}{c|}{\textbf{Performance}}  \\
    \cline{2-4}
     & \textbf{MAE$\downarrow$} & \textbf{RMSE$\downarrow$} & \textbf{$R^2\uparrow$} \\
     \hline
     \hline
     Linear Regression & 6.100 & 8.121 & 0.740\\
     \hline
     Elastic Net & 9.103 & 11.548 & 0.471\\
     \hline
     LASSO & 9.987 & 12.790 & 0.363\\
     \hline
     SVR-RBF & 5.976 & 7.875 & 0.758\\
     \hline
     Stacked-LSTM & 5.484 & 7.276 & 0.792\\
     \hline
     Temporal Attention & 5.441 & 7.239 & 0.795\\
     \hline
     PBMR-DP & \textbf{4.394} & \textbf{5.945} & \textbf{0.861} \\
     \hline
    \end{tabular}
    \caption{Different performance metrics on the Soybean Crop Yield Data performed using the published ML models.}
    \label{tab:standard_comparision}
\end{table*}

\subsection{Model Input}
\label{subsec:model_input}
The data generated explain in \cref{subsec:preprocessing} is similar to how an image is usually fed into a ConvNet as a 3D array. We will use the same ideology to directly generate (in this particular case) a 3D data array in the range 0 and 1. The data is normalized specifically to each row and not batch normalized for the entire slice. Normalization is performed since each row is sensor data over time with absolute ranges. Ex. Sensor A with a range of 0 - 100 and sensor B with a range of -1 to 25, requires different normalization. Row-based normalization will not affect the model or the output in any sense as the model is blind to how the data was generated. On testing using a batch normalization method with unique time-series data, sensors with very small ranges were found to have limited or low impact on the final results.

The generated data (\cref{fig:Data_visualzation}) is fed into the models to look for features and patterns instead of solving for the values. This approach allows us to maximize the learning ability of neural networks instead of trying to solve the best fit method. The slow trial and error of assigning a range of values to a pattern seen or observed by the model instead of solving the best equation for a set of time-based variables.

\subsection{Architecture Design}
The model relies on the feature learning/pattern recognition of CNNs. This characteristic is heavily used in classification models. The idea is to modify a few layers to convert them into a regression pattern model, which outputs a single regression yield output instead of class probability with softmax. The base architecture can be found in \cref{fig:ModelArchitecture}.

Instead of classification, we introduce an Adaptive Concat Pool layer right after the feature learning layers to understand regression data. Adaptive Concat Pool combines the Adaptive Average Pool and Adaptive Max Pooling layers defined in the PyTorch framework. This custom layer allows us to convert the problem into a FCN approach to the regression values. The use of DNNs with different optimizers and fixed hyper tuning allows us to maximize the results. These adjustments that followed the state-of-the-art architectures create a single output for each 3D input.

Bellow we describe the three architectures used in this work. As mentioned before we focused in ResNets, EfficientNets, and ResNeXt. 

\textbf{ResNet:} The addition of shortcut connections in each residual block enables gradient flow directly to the bottom layers. ResNet\cite{7780459} allows for extremely deep structures for state-of-the-art object detection performance, which is used as the baseline model for the entire approach of using 3D data in regression. Initial use case with default parameters from PyTorch models shows comparable performance and results to current solutions in the domain of Yield Estimation. The version ResNet50 was used in our experiments.

\textbf{EfficientNet:}
To demonstrate the effectiveness of scaling on both depth and resolution aspects of the existing CovNet model, a new and more mobile size baseline was designed called EfficientNet\cite{tan2020efficientnet}. The Neural Architecture was focused on optimizing the accuracy and FLOPs required to detect the same images. The base version EfficientNet b0 was used in our experiment.

\textbf{ResNeXt:}
In addition to the dimensions of depth and width of ConvNet, the paper introduces "Cardinality", a definition for the size of transformations. Allows controlling the "Network-in-Neuron" to approach optimal results in the process. Makes significant accuracy improvements on Popular ConvNets hence named as ResNeXt \cite{xie2017aggregated}. The version ResNeXt50 was used in our experiments.

\subsection{Reduced Feature Engineering}
\label{subsec:feature_engineering}
As explained in \cref{subsec:preprocessing}, the direct conversion of sensor values to the floating-point between 0 and 1 allows us full data retention. There is no approximation or wrong detection since we have no data loss during translation (normalization). Using the property of Translational invariance and Translational equivariance, we allow the models to learn from the patterns in the feature learning stage of the model. The Auto-learning ability of CNN models allows us to eliminate the need for the entire process of feature engineering, such as correlation analysis and Principal Component Analysis (PCA).


\begin{figure*}[t]
    \centering
    \includegraphics[scale=0.6]{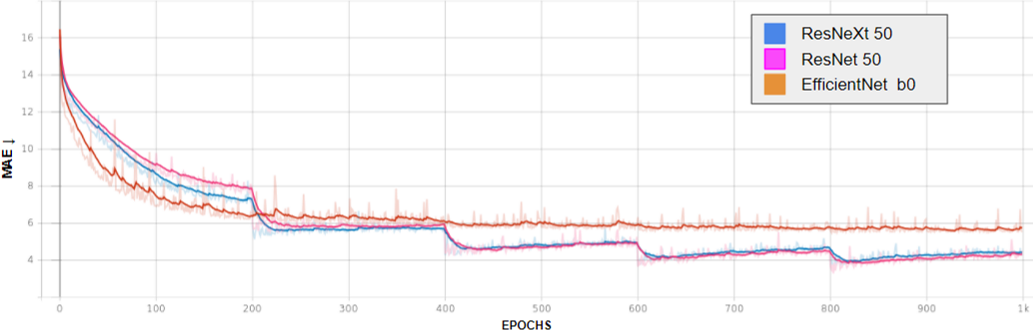}
    \caption{MAE comparison between the three samples of how the tabular input data ($x$) looks when converted to an image mapped, by our model $f(x)$ , to their predicted ($y$) value.}
    \label{fig:MAE_graph}
\end{figure*}

\section{Experiment}
\label{sec:experiment}
In the following section, the proposed data usage approach is evaluated with different state-of-the-art machine vision models. An ML tool chain was created to perform continuous tests in similar data settings and hardware setup. We conducted an ablation experiment on Crop Yield Regression Task \cite{2021}. It is a multivariable regression problem with 7 daily variables measured over a fixed time period of 214 days. The models where run in a Intel i9-10900k CPU with 128 GB 2666MHz RAM and NVIDIA  RTX 3090 with 24 GB VRAM. The data set produced image size of 214x7 which allowed to run multiple models simultaneously to produce maximum results.

Throughout the experiments, the learning rate is set to $1e^{-03}$ with a batch size of 128, 1,000 epochs and the loss after trial and error was fixed to MSEloss or L1loss. The modeling was programmed in python 3.8 using the PyTorch framework \cite{NEURIPS2019_bdbca288}. We follow \cite{xie2017aggregated,7780459,tan2020efficientnet} to construct the  Feature learning stage of the models (depth). The pooling layer is modified to a custom Adaptive Concat Layer with Fully connected layers pointed to a single output. 

\subsection{Experiments on Crop Yield Dataset}
The extensive samples of the crop yield with 93,000 samples allow the model to learn behaviors very well. The data consists of 7 weather variables, namely Average Direct Normal Irradiance (ADNI), Average Precipitation (AP), Average Relative Humidity (ARH)
Maximum Direct Normal Irradiance (MDNI), Maximum Surface Temperature (MaxSur), Minimum Surface Temperature (MinSur) and
Average Surface Temperature (AvgSur). The secondary inputs are also provided for each data point: Maturity group (MG), Genotype ID, State, Year, and Location. Each data frame points to a ground truth which is the yield.

\begin{figure*}[t]
    \centering
    \includegraphics[scale=0.6]{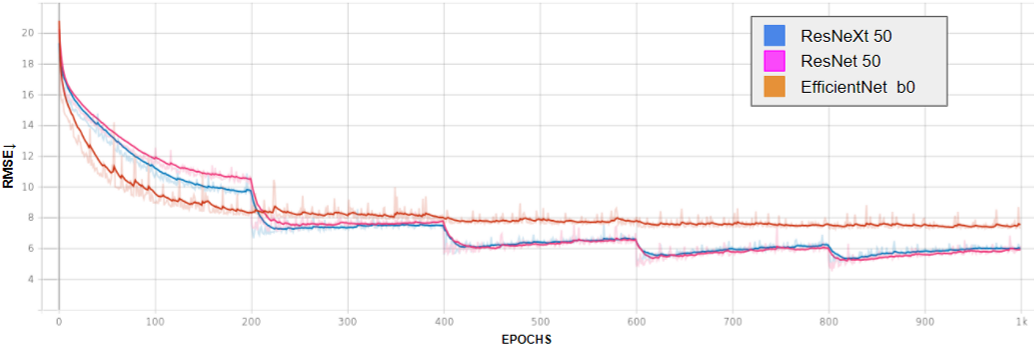}
    \caption{RMSE comparison between the three samples of how the tabular input data ($x$) looks when converted to an image mapped, by our model $f(x)$ , to their predicted ($y$) value.}
    \label{fig:RMSE_graph}
\end{figure*}

\subsection{Performance Metrics}
\label{subsec:Performance}
Unlike the accuracy metrics, which are usually associated with classification problems, to define the regression, we used the standard metrics such as Mean Average Error (MAE), Root Mean Square Error (RMSE), and $R^2$ to evaluate the performance. The loss function used in the model is MSEloss or L1loss in the PyTorch framework. 
k-cross-validation is performed to overcome over-fitting of data. Significant improvements are noted in validation datasets. Significant improvements are noted in validation datasets.  The data was tested and compared with the same test dataset as the MLCAS2021 competition to keep the results and metrics constant and form a common comparison baseline.

\begin{figure*}[t]
    \centering
    \includegraphics[scale=0.6]{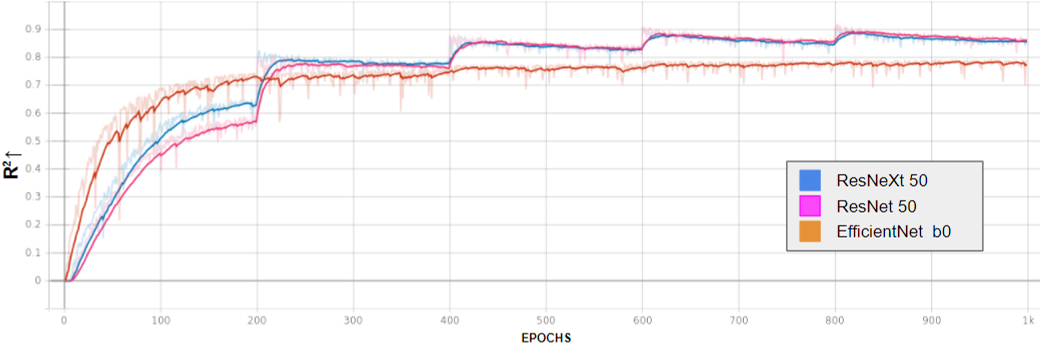}
    \caption{$R^2$ comparison between the three samples of how the tabular input data ($x$) looks when converted to an image mapped, by our model $f(x)$ , to their predicted ($y$) value.}
    \label{fig:R2_graph}
\end{figure*}

Figures \ref{fig:MAE_graph}-\ref{fig:R2_graph} show the performance metrics of the top three models conducted on the crop yield data set with the proposed architecture.
In \Cref{fig:MAE_graph}, we see that Efficient Net b0 as designed learns faster, but as the model is not deep enough, it saturates after 400 epochs. Both ResNet and ResNeXt learn slower but restarts the learning process at each k-fold change.

\section{Results and Discussion}
\label{sec:results}

\textbf{Comparison with different models:} \Cref{tab:CNN_Multi_model_metric} shows the results gathered when comparing the different networks with different optimizers. Here we explore Stochastic Gradient Descent, Adam Optimizer, and LARS with the same parameters and metrics described in 4. We found that ResNeXt50 with SGD optimizer performed the best in the three different metrics used for this experiment. The second and third best models were ResNet50 with SGD and LARS, respectively. This tells us that for this use case, having an SGD is better during the training process of our network.

\textbf{Comparing Competition approaches:} \Cref{tab:competiion_metrics} shows the performance of different online teams from the MCLAS Challenge. The best models were shown in the online leaderboard and available publicly for the challenge. Some of these works relied upon heavy statistical analysis and feature engineering in multiplying the number of available features to improve learning parameters for the data. Most of the results involved using ensemble techniques to combine weights generated using different models to get the best results. Our approach is simpler with just the modified DNNs to become a regression model with a custom data loader to convert Real-time data into an image type array. This table shows that our model outperforms the methods in the competition except for one method. We are able to outperform QU(exp006) only in MAE but not in the other metrics. It is noteworthy that we have trained our model without optimizing the hyperparameters as we wanted our solution to work as a general method. Fine tuning hyperparameters would help improve our results.

\textbf{Comparison with state-of-the-art results:} \Cref{tab:standard_comparision} shows the crop yield prediction dataset results. Our results prove a dramatic increase in prediction performance with a simple change in how data is used. In addition, our model approach allows for faster data to model regression without the need for analysis of the correlation between the inputs and the output. This table shows the different published works that used our same dataset. We can see that our model outperforms these methods in each selected metrics. 

\section{Conclusion}
\label{sec:conclusion}
This work provides a pattern-based approach for multivariable regression. With our sensor-to-image conversion, we are able to bring computer vision and convolutional neural network techniques to regression tasks. Our method of sensor-to-image conversion is completely lossless. Our experiment with multiple models and different optimizers proves the validity of our method. We have outperformed every classical approach and are at par with the best ensemble methods. In addition, we hope to make a significant impact with tabular data and advance the research even further in these areas.

{\small
\bibliographystyle{ieee_fullname}
\bibliography{references}
}

\end{document}